\documentclass[10pt,twocolumn,letterpaper]{article}

\usepackage[T1]{fontenc}
\usepackage{times}
\usepackage[hmargin=0.72in,top=0.8in,bottom=0.85in,columnsep=0.28in]{geometry}
\usepackage{microtype}
\usepackage{amsmath,amssymb,graphicx,booktabs,xspace}
\usepackage[font=small,labelfont=bf,labelsep=period]{caption}
\usepackage[numbers,sort&compress]{natbib}
\usepackage{xcolor}
\definecolor{linkblue}{HTML}{24527B}
\usepackage[breaklinks,colorlinks,allcolors=linkblue]{hyperref}
\usepackage[nameinlink,noabbrev]{cleveref}
\hypersetup{
  pdftitle={Learning Control-Oriented World Models for Fluid–Solid Interaction},
  pdfauthor={Kangrui Zhang, Runze Zhang, Ruihong Cen, Yuchen He, Bo Ren}
}

\newcommand{\ours}{FSI-WM\xspace}
\newcommand{\Hlen}{L}
\makeatletter
\DeclareRobustCommand{\etal}{\textit{et al}\@ifnextchar.{}{.\@\xspace}}
\makeatother

\title{\bfseries Learning Control-Oriented World Models for\\Fluid--Solid Interaction}
\author{Kangrui Zhang\textsuperscript{1,*} \quad Runze Zhang\textsuperscript{1,*} \quad
Ruihong Cen\textsuperscript{2,*} \quad Yuchen He\textsuperscript{1} \quad Bo Ren\textsuperscript{1}\\[0.45em]
\normalsize \textsuperscript{1}Nankai University \qquad
\normalsize \textsuperscript{2}National University of Singapore}
\date{}

\begin{document}
\twocolumn[{%
  \begin{@twocolumnfalse}
    \maketitle
    \begin{minipage}{\textwidth}
      \captionsetup{type=figure}
\centering
  \includegraphics[width=\textwidth]{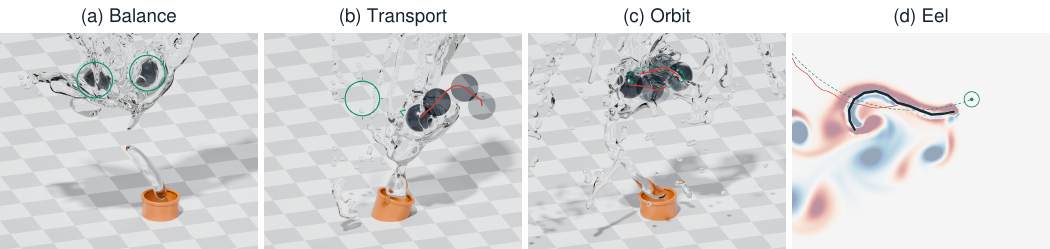}
  \caption{\textbf{Four fluid--solid control tasks learned with \ours.} (a)--(c) A nozzle moves balls only through its water jet (3D MPM): balancing two balls at fixed targets, transporting a ball to a sequence of goals, and moving two balls along a horizontal orbit. (d) An articulated eel follows a reference path by pushing against the surrounding fluid (2D lattice-Boltzmann; color: vorticity). Green markers show task targets; translucent balls show recorded past positions. In every task, the world model predicts the bodies and the actuator but not the flow field, and is learned online together with the policy.}
  \label{fig:overview}

    \end{minipage}\par\vspace{1.2em}
  \end{@twocolumnfalse}
}]
\begingroup
  \renewcommand{\thefootnote}{\fnsymbol{footnote}}
  \begin{NoHyper}
    \footnotetext[1]{These authors contributed equally to this work.}
  \end{NoHyper}
\endgroup
\begin{abstract}

Learning control policies for coupled fluid--solid systems is important for a wide range of applications, but remains computationally expensive. Model-free reinforcement learning requires extensive simulator interactions, while differentiable simulation requires solver-specific derivatives, introducing implementation challenges and computational overhead. We introduce \ours, an online model-based reinforcement learning framework that jointly learns a compact differentiable world model and a control policy from black-box simulator interactions. Rather than learning detailed fluid--solid dynamics, \ours captures low-dimensional dynamics specific to policy optimization. Policies are optimized by backpropagating returns through short imagined rollouts, using closed-form task rewards, smooth surrogates for discrete events, and a learned critic to account for longer-term returns. We evaluate \ours on four control tasks across MPM and lattice-Boltzmann simulators in 2D and 3D. Given the same observations, \ours learns policies with higher returns than the state-of-the-art baseline on all four tasks and, on the jet-driven tasks, reaches the baseline's final return with several times fewer simulator transitions.

\end{abstract}

\section{Introduction}
\label{sec:intro}

Learning control policies for coupled fluid--solid systems is essential for tasks such as jet-driven manipulation and aquatic locomotion. However, the computational cost of fluid--solid simulation makes policy optimization expensive. Model-free reinforcement learning methods~\cite{ma2018fluid,ren2022versatile,chen2025fast} require extensive simulator interactions to learn effective policies. Differentiable simulation enables gradient-based optimization through differentiable programming~\cite{hu2020difftaichi}, SPH-based coupling~\cite{li2023difffr,cao2026e2esph}, flow-map adjoints~\cite{li2025adjoint}, and differentiable multiphysics simulation~\cite{xing2025stabilizing}. Yet computing these gradients requires differentiating the solver's time-stepping operations and fluid--solid coupling, introducing solver-specific implementation requirements and additional computational costs. These limitations motivate an approach that reduces simulation demands while remaining independent of solver derivatives.

We introduce \ours, an online model-based reinforcement learning framework for fluid--solid control. Our key idea is to learn a control-oriented world model from black-box simulator interactions and use it to compute policy gradients. Unlike conventional learning-based fluid models designed to learn detailed flow fields~\cite{wang2026worldparticle}, our control-oriented world model targets low-dimensional dynamics sufficient for policy optimization. The model is trained alongside the policy, which is optimized by backpropagating returns through short imagined rollouts. We evaluate task rewards in closed form on predicted states and replace discrete events, such as falls and goal deadlines, with differentiable surrogates. This enables policy optimization through a control-oriented model without differentiating the underlying solver.

We evaluate the same learning algorithm on four tasks across two solvers: jet-driven balancing, transport, and orbit tracking in a 3D MPM fluid, and path following with a 12-link eel in a 2D lattice-Boltzmann fluid (\cref{fig:overview}). Using the same observations as the state-of-the-art baseline~\cite{chen2025fast}, \ours{} attains higher final returns on all four tasks; on the jet-driven tasks it reaches the baseline's final return with 7--10$\times$ fewer simulator transitions, and on the eel it succeeds on all 25 test paths at 2.4M transitions, whereas the baseline succeeds on at most about half.

\noindent Our contributions are:
\begin{itemize}
    \item An online model-based framework for fluid--solid control that learns compact, control-oriented dynamics from black-box simulator interactions without requiring a differentiable solver.
    \item A differentiable rollout formulation combining learned dynamics, closed-form task rewards, and smooth event surrogates for policy optimization.
    \item An evaluation across four tasks and two fluid solvers, showing substantial gains over the state-of-the-art baseline in sample efficiency and final performance.
\end{itemize}

\section{Related Work}
\label{sec:related}

\paragraph{Learning control of fluid--solid interaction.}
Ma~\etal~\cite{ma2018fluid} trained deep RL controllers that move 2D fluid jets to balance, juggle, or steer rigid bodies, and showed that the fluid velocity field is a useful policy input.
Ren~\etal~\cite{ren2022versatile} extended this line to 3D with meta-RL and hindsight relabeling for multi-task control, using an autoencoder to compress the velocity field.
Chen~\etal~\cite{chen2025fast} improved stability and convergence with a twin-delayed actor--critic, a Boltzmann softmax operator, a two-step value estimator, and FEDG for guided exploration.
In fluid mechanics, deep RL has been applied to active flow control~\cite{rabault2019artificial}, and model ensembles have been used to generate extra training data for model-free policy optimization~\cite{weiner2025model}.
All of these controllers obtain their policy updates from sampled rewards.
We instead learn the policy through a world model of the dynamics that matter for control, and compare against the state-of-the-art method of Chen~\etal with the same inputs.

\paragraph{Learned models of fluids.}
Learned particle simulators such as DPI-Net~\cite{li2019learning} predict fluid and rigid particles and support trajectory optimization for manipulation, but they are trained offline and model every particle.
In SoftGym~\cite{lin2020softgym}, a latent world model learned from images performs poorly on liquid pouring and transport, which illustrates how hard liquid dynamics are to learn.
CFC~\cite{dou2025cfc} learns a two-level world model from SPH data, combining a physics-informed fluid model with a character model through predicted forces, and trains a character policy with gradients through it.
There, the character moves through the fluid, which acts as its environment.
In our tasks the fluid is the means of actuation.
These models aim to reproduce the physics: every particle, the flow field, or the forces between fluid and body. Our world model is control-oriented instead: it predicts only the bodies and the actuator, and is learned online from the policy's own interaction with a black-box simulator, so it is accurate where the policy goes.

\paragraph{Differentiable simulation for control.}
Differentiable programming frameworks such as DiffTaichi~\cite{hu2020difftaichi} and FluidLab~\cite{xian2023fluidlab} enable gradient-based control of soft bodies and fluids.
Differentiable fluid--rigid coupling has been realized via SPH~\cite{li2023difffr,cao2026e2esph,schenck2018spnets} and adjoint methods on flow maps~\cite{li2025adjoint}.
Analytic policy gradients through parallel differentiable simulators accelerate policy learning~\cite{xu2022accelerated}, but gradients through stiff contact and chaotic dynamics can have high variance or bias~\cite{suh2022differentiable}.
Rewarped~\cite{xing2025stabilizing} brings first-order RL to differentiable MPM in Warp, but couples rigid bodies to the material only one way: kinematic rigid bodies move the particles, and the particles never push back.
Jet-driven control requires the opposite direction, since the fluid is what moves the solids, and a nozzle that emits a discrete number of particles provides no gradient with respect to its flow rate.
Our approach needs neither: it treats the coupled simulator as a black box and obtains gradients from a learned model of the control-relevant dynamics instead.

\paragraph{Model-based RL and world models.}
Learned dynamics models have long been used to improve sample efficiency~\cite{sutton1991dyna,deisenroth2011pilco,ha2018world}.
MBPO~\cite{janner2019trust} generates short model rollouts for a model-free learner, while SVG~\cite{heess2015learning}, MAAC~\cite{clavera2020model}, and Dreamer~\cite{hafner2020dream,hafner2023mastering} backpropagate value gradients through learned models.
TD-MPC~\cite{hansen2022temporal,hansen2024tdmpc2} combines latent models with planning.
SuperMimic~\cite{liao2026supermimic} showed that a single world model of a humanoid's state, trained online together with the policy, trains controllers with orders of magnitude less compute than model-free RL.
Our policy optimization follows this design.
There, the predicted state is the full state of a rigid body; in fluid--solid interaction, most of the state is the fluid, and the question becomes which part of the dynamics the model should represent. We answer it for jet-driven and swimming tasks and evaluate against the state-of-the-art controller.
Partially observed control is commonly handled with stacked observations~\cite{mnih2015human} or recurrent models~\cite{hausknecht2015deep,ni2022recurrent}, and our world model can likewise read a short causal history.

\section{Method}
\label{sec:method}

\begin{figure*}[t]
  \centering
  \includegraphics[width=\textwidth]{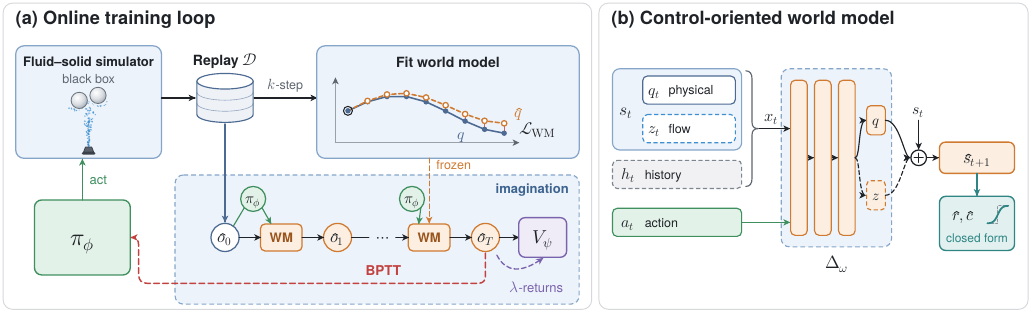}
  \caption{\textbf{Overview of \ours.} (a) The world model, the policy $\pi_\phi$, and the critic $V_\psi$ are trained from scratch in one online loop: transitions from a black-box fluid--solid simulator, MPM or lattice-Boltzmann in our benchmarks, fill the replay, the world model is fitted on $k$-step windows, and the policy is improved by backpropagating through imagined rollouts from replay states, with the critic fitted to $\lambda$-returns. (b) The world model $\Delta_\omega$ maps the state $s_t$ (physical state $q_t$ and optional fluid observation $z_t$), the optional history $h_t$, and the action $a_t$ to the next state; reward and survival are evaluated in closed form.}
  \label{fig:method}
\end{figure*}

\subsection{Problem formulation}
\label{sec:problem}
We consider a scene with a fluid, one or several solid bodies, and an actuator that the controller drives.
The actuator may be a moving water nozzle or the joints of a swimming body.
The controller influences the bodies only through the fluid.
The simulator advances the full state, including tens of thousands of fluid particles or lattice cells, with several substeps per control step.
The controller does not observe this state.
At control step $t$ it observes
\begin{equation}
  o_t = \big(q_t,\; z_t,\; g_t\big),
  \label{eq:obs}
\end{equation}
where $q_t$ is the physical state, $z_t$ an optional low-dimensional observation of the fluid, and $g_t$ the task variables.
The physical state contains the positions and velocities of the bodies and the state of the actuator.
On the jet tasks, $z_t$ is a flow code produced by a frozen autoencoder of the grid velocity field, as in prior work~\cite{ren2022versatile,chen2025fast}. On the eel, it is a lateral line: the fluid velocity sampled at 24 points, three cells to either side of the centres of the 12 body segments, and expressed in the body frame, analogous to the lateral-line organ of fish. It is local and needs no learned encoder.
The task variables are, for example, the current goal and the elapsed time, a moving target, or the path ahead.
Which of these quantities are available is defined by each benchmark, and we use the same observation for our method and for the baselines (\cref{sec:setup}).
The action $a_t\in[-1,1]^{d_a}$ sets the accelerations, flow rate, or joint targets of the actuator.

The reward $r_{t+1}$ and the termination $d_{t+1}\in\{0,1\}$ are known functions of $q_{t+1}$ and the task variables.
For example, an episode terminates when a ball falls below a height threshold or misses a goal window, or when a swimmer leaves its path.
With the survival weight $c_0{=}1$, $c_{t+1}{=}c_t(1-d_{t+1})$, we seek a policy $\pi_\phi$ that maximizes
\begin{equation}
  J(\phi)=\mathbb{E}\Big[\textstyle\sum_{t\ge 0}\gamma^t c_t\, r_{t+1}\Big]
  \label{eq:objective-true}
\end{equation}
with as few simulator transitions as possible, since each transition requires a full fluid--solid simulation step.
The observation $o_t$ does not determine the future of the system: the water that will act on a body in the next steps is not part of $q_t$, and $z_t$ compresses it lossily.

\subsection{A control-oriented world model}
\label{sec:wm}
The simulator provides no gradients, and differentiating an MPM or lattice-Boltzmann solver through long fluid--solid rollouts is costly in memory and numerically fragile~\cite{suh2022differentiable}.
We therefore learn a world model that supplies predictions and gradients for policy optimization (\cref{sec:policy}).
Its design follows one principle: represent the dynamics that the controller needs, not the full physics.
The reward and the termination depend only on the bodies, and the policy acts only through the actuator, so the model predicts these quantities and treats the fluid as a hidden cause of their motion.
This turns a simulation with tens of thousands of particles or cells into a prediction problem over a few dozen dimensions, as continuum mechanics replaces molecular dynamics with a description adequate for the question at hand.

\paragraph{What the model predicts.}
Let $s_t=(q_t,z_t)$ be the part of the observation that the model predicts: the physical state and, when the benchmark provides it, the fluid observation (the flow code or the lateral line).
An MLP $\Delta_\omega$ takes the current state, the optional state--action history $h_t$ (\cref{sec:inputs}), and the action $a_t$, and predicts the change of the state over one control step:
\begin{equation}
  \hat s_{t+1} = s_t + \Delta_\omega(x_t, a_t), \qquad x_t = (s_t, h_t).
  \label{eq:dynamics}
\end{equation}
$\Delta_\omega$ is a single MLP whose output layer predicts the change of every component of $s_t$.
Its inputs are normalized by statistics of the real states and its outputs are residuals scaled by statistics of the real one-step changes; these are running statistics on the MPM tasks and are fitted once before training on the eel.
State components that are known to stay constant, such as the fixed height of a nozzle, are excluded from the residual.
On the eel, the planar pose of the body is advanced by composing the predicted rigid displacement with the current pose.
The model may also leave the fluid observation out of $s_t$; the policy then keeps the last real fluid observation throughout the rollout.

\paragraph{Closed-form task terms.}
The reward and the termination are known functions of $\hat q$.
We evaluate them in closed form on the predictions, so the model needs no reward or termination head and no labels of rare failure events.
This focuses the learned part of the model on the physics: the response of the bodies and actuator to the fluid and the action.

\paragraph{Model learning.}
We fit $\omega$ on windows of $k$ consecutive replay transitions ($k{=}8$ on the MPM tasks and 32 on the eel).
Each window is seeded at a logged state and rolled forward autoregressively on the logged actions:
\begin{equation}
  \mathcal{L}_{\mathrm{WM}}(\omega) = \mathbb{E}\Big[\frac{1}{k}\sum_{j=1}^{k} \sum_{\mathcal{G}} \frac{1}{|\mathcal{G}|}\sum_{i\in\mathcal{G}} \Big(\frac{\hat s_{j,i} - s_{j,i}}{\sigma_i}\Big)^{2}\Big],
  \label{eq:wmloss}
\end{equation}
where $\mathcal{G}$ ranges over groups of state components (body positions, body velocities, actuator, and the fluid observation when it is predicted) and $\sigma_i$ is the standard deviation of component $i$ in the real data.
Averaging within each group keeps the much larger fluid observation and the easily predicted actuator from dominating the bodies.
Windows never cross episode boundaries.
Because the model consumes its own predictions, the multi-step loss trains it to correct the errors that it encounters in imagination.
On the MPM tasks, this keeps rollouts of $T{=}16$ steps stable although $k{=}8$.

\paragraph{Smooth task events.}
Unrolling \cref{eq:dynamics} with a policy produces a predicted trajectory that is differentiable with respect to every action.
Its terminations, however, are discrete events, such as a ball that either falls or does not, or a goal cell that is either reached within its window or missed.
Hard indicators of such events carry no gradient, so a policy that does not yet cause an event in imagination cannot learn to avoid it.
Since each event is a known threshold on the predicted state, we replace the threshold by a sigmoid of the signed distance to it.
For a fall below height $y_f$ and for reaching a goal cell of size $b$,
\begin{align}
  \hat d^{\,\mathrm{fall}}_{t} &= \sigma\!\big((y_f - \hat y_t)/w\big), \nonumber\\
  \hat e^{\,\mathrm{goal}}_{t} &= \textstyle\prod_{\ell}\sigma\!\big((\tfrac{1}{2}b_\ell - |\hat p_{t,\ell}-g_{t,\ell}|)/w\big),
  \label{eq:events}
\end{align}
with a width $w{=}2$\,mm much smaller than the bodies.
The soft goal indicator accumulates into a reached flag that is checked at the end of each goal window.
Several events combine multiplicatively into a differentiable survival weight $\hat c_t$.
The surrogates coincide with the true events away from the thresholds.
The simulator and all evaluations use the exact events.
The smoothed fall indicator also replaces the hard one in the fall penalty of the reward.
On the eel we keep the terminations hard, so reaching the end of the path or leaving it carries no gradient; the policy is steered there by the smooth progress, cross-track, and heading terms of the reward.

\subsection{World-model-based policy optimization}
\label{sec:policy}
We optimize the policy in imagination with the world model held fixed.
This step follows the structure of SuperMimic~\cite{liao2026supermimic}: first-order policy gradients by backpropagation through time (BPTT) over a short horizon, and a critic fitted to imagined $\lambda$-returns.
What is specific to our setting is that the imagined trajectory consists of predicted solids and actuator, its rewards and events are the closed-form terms of \cref{sec:wm}, and the policy acts on the benchmark observation reconstructed from the prediction.

\paragraph{Imagined rollouts.}
Each rollout starts from a state sampled uniformly from the replay buffer, which keeps the most recent transitions.
At imagined step $t$, the policy acts on the predicted observation, $a_t=\pi_\phi(\hat o_t)$.
The world model predicts $\hat q_{t+1}$ and $\hat z_{t+1}$, and the reward and event surrogates are evaluated on the prediction.
Task variables evolve in imagination as in the real task: the clock advances, and when a goal window ends with the goal reached, a new goal is drawn from its distribution without using the real future.

\paragraph{Actor optimization.}
The policy maximizes the imagined return over $T$ steps ($T{=}16$, or 32 on the eel),
\begin{equation}
  \hat J(\phi) = \mathbb{E}_{s_0}\Big[\sum_{t=0}^{T-1}\gamma^t \hat c_t\, \hat r_{t+1} + \gamma^{T}\hat c_T\, V_\psi(\hat o_T)\Big].
  \label{eq:objective}
\end{equation}
Since all terms are differentiable, BPTT through the frozen world model yields the analytic gradient $\nabla_\phi \hat J$ along the predicted trajectory of the bodies.
Model-free methods~\cite{ma2018fluid,ren2022versatile,chen2025fast} instead learn a state--action critic from sampled rewards and improve the policy only through that critic.
Our gradient differentiates the return with respect to each action through the learned physics.

\paragraph{Value learning.}
The critic $V_\psi$ estimates the return beyond the imagination horizon.
It is fit to $\lambda$-returns computed backward over the imagined transitions,
\begin{align}
  \hat R_t &= \hat r_{t+1} + \gamma(1-\hat d_{t+1})\big[(1-\lambda)V_{t+1} + \lambda \hat R_{t+1}\big], \nonumber\\
  \mathcal{L}_V(\psi) &= \mathbb{E}\Big[\textstyle\sum_{t=0}^{T-1}\big(V_t - \hat R_t\big)^2\Big],
  \label{eq:critic}
\end{align}
with $V_t=V_\psi(\hat o_t)$, starting from $\hat R_T=V_T$.
The targets are computed once per rollout, using a snapshot of the critic, and held fixed while $\psi$ is fit.
The critic is essential for fluid--solid control.
A dropped ball ends a balancing episode that would otherwise last 8{,}000 steps, so most of the return lies far beyond any affordable horizon.

\paragraph{Online schedule.}
Interaction and learning are interleaved, in the Dyna tradition~\cite{sutton1991dyna}.
Every 64 simulator transitions, collected with Gaussian exploration noise, the world model, the policy, and the critic are updated in this order.
None of them is ever frozen, so the world model is refit as the policy reaches new regions of the state space. The model therefore does not need to be accurate everywhere, only on the states that the current policy visits; this is the second sense in which it is control-oriented.
Transitions collected by earlier policies remain valid for model fitting, because the physics does not depend on the policy that chose the actions.
All three networks start from random initialization.
The frozen autoencoder behind the flow code on the jet tasks is part of the benchmark, shared with the baselines.

\subsection{Task-dependent inputs}
\label{sec:inputs}
The learning algorithm above is shared by all tasks.
What varies is the input $x_t$ of the world model. Because the model does not simulate the fluid, it must infer the fluid's effect on the bodies from what it reads, and the input is extended when the benchmark observation does not suffice to predict the next physical state.

\paragraph{State--action history.}
When the observation misses fluid that is in flight or forces that are visible only over time, the world model can be conditioned on a causal window of the $\Hlen$ most recent physical states and executed actions,
\begin{equation}
  h_t = \big((q_{t-i},\, a_{t-i},\, m_{t-i})\big)_{i=1}^{\Hlen},
  \label{eq:history}
\end{equation}
where $m$ marks entries that are valid since the last reset.
The window is not predicted.
It is shifted after every step by appending the newest predicted physical state and the executed action, which is the policy's action in imagination and the logged action in model learning.
Only the window at the start of a rollout holds real states, so real future quantities never enter it.
When the input of all methods includes this history (\cref{sec:setup}), the policy and value read it as well.
In a jet task, the water that will strike a ball over the next steps was emitted at nozzle poses and flow rates that $h_t$ records.

\section{Experiments}
\label{sec:experiments}

\subsection{Benchmarks and setup}
\label{sec:setup}
We evaluate on four fluid--solid control tasks that differ in how the fluid acts on the solid.
In Balance, Transport, and Orbit, a moving nozzle controls balls only through its water jet (3D MLS-MPM~\cite{hu2018moving} in Warp~\cite{macklin2022warp}), so the effect of an action arrives with a delay and must be anticipated.
Balance keeps two balls aloft for up to 8{,}000 steps~\cite{chen2025fast}, Transport carries one ball through a random sequence of goal cells with a deadline for each, and Orbit makes two balls follow moving targets.
The eel moves only by pushing against the surrounding fluid (2D lattice-Boltzmann), so its motion results entirely from the fluid's reaction to its own body. It must follow a reference path: an episode succeeds when the eel reaches the end of the path within a small lateral distance of it, and fails when its lateral deviation or its heading error relative to the path tangent exceeds a threshold.
Together they cover two fluid solvers, 2D and 3D scenes, and nozzle and joint actuation.

\paragraph{Baselines and inputs.}
We compare to Chen~\etal~\cite{chen2025fast}, the state-of-the-art controller for fluid--solid interaction, and to SAC~\cite{haarnoja2018soft,ren2022versatile}.
Chen~\etal train a twin actor--critic whose critics are fitted to two-step Bellman targets with a Boltzmann softmax over sampled target actions, and optionally add guided exploration (FEDG), which requires an auxiliary reward designed by hand for each task.
We first ran the authors' public Taichi code with its original hyperparameters; across runs on several machines, the variants with and without FEDG performed comparably on Balance.
We then ported the code from a single Taichi environment to four parallel environments in Warp, which accelerates the baselines as much as our method and makes the baseline about 4$\times$ faster per transition than reported by Chen~\etal on the same GPU type.
In the port, the two were again comparable, but FEDG was about 1.7$\times$ slower per transition, because every update also trains on a batch from the auxiliary-task replay; we therefore compare to the variant without FEDG.
All methods read the same input.

\paragraph{Protocol.}
All methods train from random initialization. On the eel, both methods also use a fixed set of scripted-gait clips that is not counted on the x-axis: the baseline starts half of its training episodes from swimming states taken from them, and our world model is fitted partly on them, starts half of its imagined rollouts from them, and receives 500 model updates on them before the policy is trained.
Every 25{,}000 real transitions we run 10 deterministic evaluation episodes in a separate simulator, which does not count toward the budget. Curves show the mean return smoothed by a 5-point moving average, with bands of one standard deviation over the evaluation episodes. On the eel, we evaluate on its 25 test paths every 200{,}000 transitions.
We report cost both in real transitions and in training GPU-hours on a single GPU, which exclude the time spent on evaluation.
One hyperparameter set is shared by the three MPM tasks; the eel uses larger batches, windows, and horizons.
All results use a single training seed; however, the advantage of our method is readily reproduced with other seeds.
Benchmark and implementation details, numerical summaries, training costs, and the FEDG comparison are in Appendices~\ref{sec:suppl-details}--\ref{sec:suppl-results}.

\subsection{Comparison with baselines}
\label{sec:main}
\label{sec:systems}
\begin{figure}[t]
  \centering
  \includegraphics[width=\linewidth]{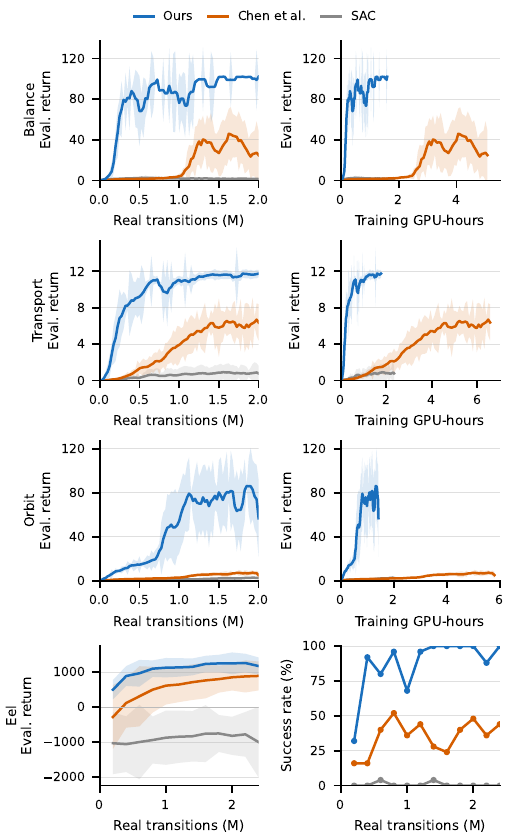}
  \caption{\textbf{Comparison with baselines.} Rows: Balance, Transport, Orbit, eel. Left: return vs.\ real transitions. Right: return vs.\ training GPU-hours for the jet tasks, and success rate on the 25 test paths for the eel (not smoothed). Eel: up to 2.4M transitions. Within each row, all methods read the same input.}
  \label{fig:main}
\end{figure}

\Cref{fig:main} compares \ours with the baselines.

\begin{figure}[t]
  \centering
  \includegraphics[width=\linewidth]{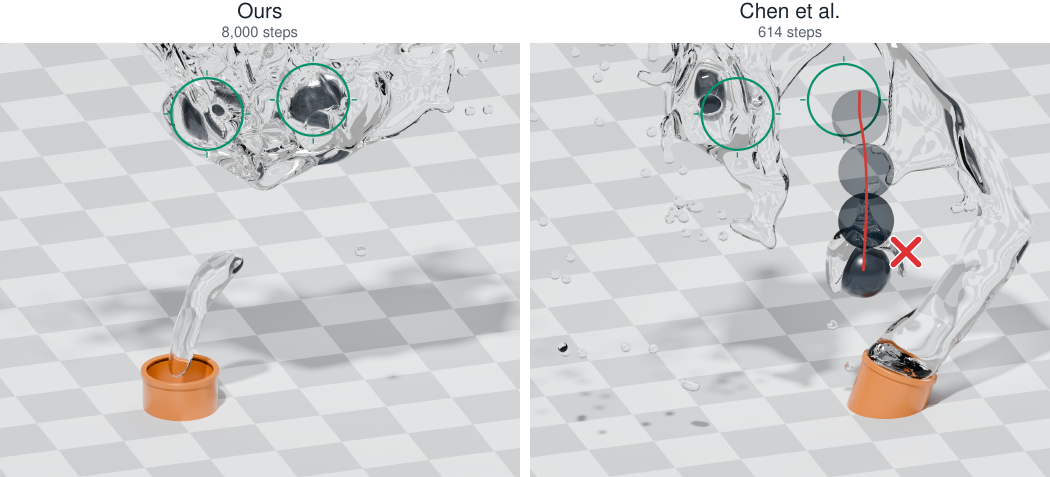}
  \caption{\textbf{Balance.} Left: ours keeps both balls aloft at step 8{,}000. Right: Chen~\etal at step 614, just before a ball falls (red cross); gray copies and red line: its preceding motion. Green circles mark the targets of both balls.}
  \label{fig:balance-qualitative}
\end{figure}

\paragraph{Balance.}
\ours keeps both balls aloft for the full 8{,}000 steps in all 10 evaluation episodes from 0.3M transitions on (\cref{fig:balance-qualitative}).
It reaches the final return of Chen~\etal after 0.25M transitions and ends with 4.1$\times$ that return, whereas SAC does not learn the task at all.

\paragraph{Transport.}
Here the ball must reach a sequence of goal cells before their deadlines.
\ours reaches the baseline's final return after 0.3M transitions, 7$\times$ fewer than the baseline's 2M, and ends with 1.8$\times$ its final return.
SAC does not learn the task within 2M transitions.

\paragraph{Orbit.}
The balls must follow targets that move around a circle, so the controller has to keep them in motion rather than hold them in place.
\ours reaches the baseline's final return within 0.2M transitions, and its final return is 11$\times$ the baseline's.
Its late-training evaluations fluctuate considerably, yet stay far above both baselines; SAC does not learn the task.

\begin{figure}[t]
  \centering
  \includegraphics[width=\linewidth]{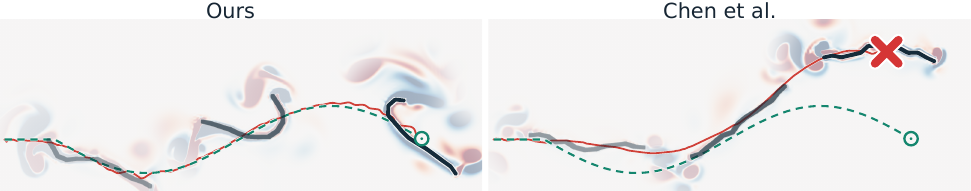}
  \caption{\textbf{Eel on the same test path.} Green dashes: reference; red: trajectory; red cross: the baseline exceeds the cross-track limit.}
  \label{fig:eel-qualitative}
\end{figure}

\paragraph{Eel.}
The eel uses a different solver and a joint-space action, and senses the fluid only through its lateral line.
Our return is above the baseline's at every evaluation, and the gap is larger in success rate: our controller succeeds on 92\% of the 25 test paths after 0.4M transitions and on all of them from 1.4M to 2.0M transitions, whereas the baseline never exceeds 52\% (see \cref{fig:eel-qualitative} for an example). Its training time on the eel is similar to the baseline's.

\subsection{Ablations}
\label{sec:ablation}
\begin{figure}[t]
  \centering
  \includegraphics[width=\linewidth]{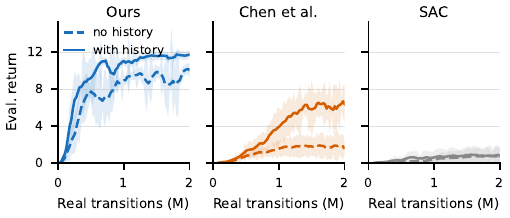}
  \caption{\textbf{History ablation on Transport.} Each method with (solid) and without (dashed) the last 8 states and actions.}
  \label{fig:history}
\end{figure}

\paragraph{History.}
Prior learned controllers for fluid--solid interaction~\cite{ma2018fluid,ren2022versatile,chen2025fast} act on the current state and fluid features only. We find that adding the last 8 states and actions to the input improves the baseline as well as ours.
\Cref{fig:history} shows Transport, where all variants already read the flow code: with history, the final return of Chen~\etal rises about fourfold, from 1.6 to 6.4, and ours rises from 10.1 to 11.8 with markedly more stable training; SAC does not learn the task either way.
Our intuition is that the recent states and actions implicitly carry information about the flow, which complements the flow code.
In \cref{sec:main}, all methods read the history on Transport and Orbit.

\begin{figure}[t]
  \centering
  \includegraphics[width=\linewidth]{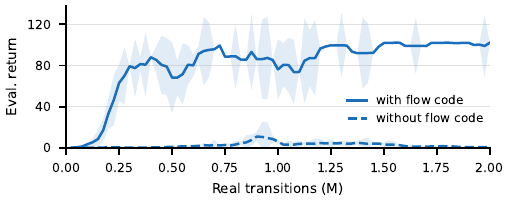}
  \caption{\textbf{Flow-code ablation on Balance.} Our world model with and without the flow code; neither variant reads the history.}
  \label{fig:balance-abl}
\end{figure}

\paragraph{Flow code.}
Our world model never simulates the fluid, but it must know enough about it to predict the bodies.
On Balance, without history input, a world model that neither reads nor predicts the flow code does not learn the task (\cref{fig:balance-abl}): the balls hover on the jet, and how the water will hit them depends on the flow around them, which the current physical state alone does not determine.

\begin{figure}[t]
  \centering
  \includegraphics[width=\linewidth]{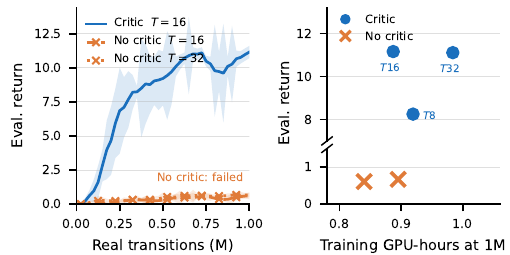}
  \caption{\textbf{Critic ablation on Transport.} \textit{Left}: learning curves. Without a critic the policy fails at every tested horizon, none of which covers the 250-step goal window; the critic supplies the return beyond it. \textit{Right}: return vs.\ training GPU-hours at 1M transitions.}
  \label{fig:critic}
\end{figure}

\paragraph{Is the critic necessary?}
Without a critic, the policy fails on Transport at every horizon we test (\cref{fig:critic}, left).
A goal window lasts 250 steps, far longer than any tested imagination horizon, so without a value estimate beyond $T$ most imagined rollouts do not reach the outcome of a goal.
With the critic, $T{=}16$ gives the best return per GPU-hour (\cref{fig:critic}, right), and we use it throughout.

\begin{figure*}[t]
  \centering
  \includegraphics[width=\textwidth,trim=0 270 0 0,clip]{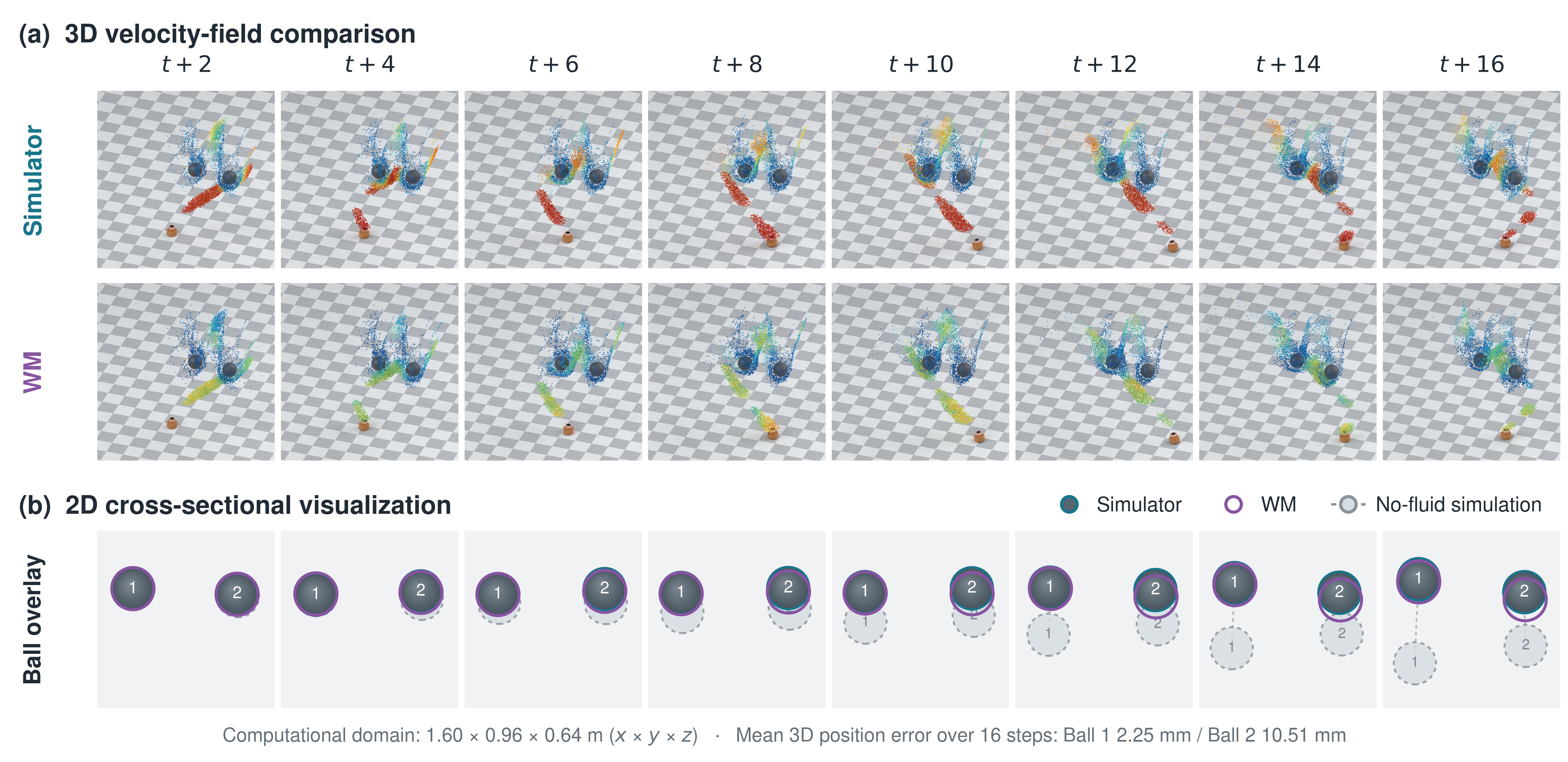}
  \caption{\textbf{World-model visualization on Balance.} 16-step open-loop prediction from the same state and actions. (a) Particles colored by speed on a scale shared by both rows; WM: flow decoded from the predicted flow code. (b) Balls of the simulator, the WM, and a simulation without fluid. The WM captures control-relevant dynamics in a compact, low-dimensional representation, closely predicting ball motion without reproducing exact velocity magnitudes or fine-scale splashing.}
  \label{fig:wm-vis}
\end{figure*}

\subsection{World-model analysis}
\label{sec:analysis}
\begin{figure}[t]
  \centering
  \includegraphics[width=\linewidth]{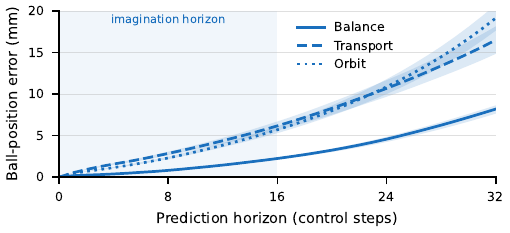}
  \caption{\textbf{World-model accuracy.} Root-mean-square error of the predicted ball coordinates, rolled out on recorded actions from evaluation episodes, against the prediction horizon (one step is 3.8\,ms). Bands: 95\% bootstrap intervals over episodes. Shaded: the imagination horizon $T{=}16$. The ball radius is 50\,mm.}
  \label{fig:wm-analysis}
\end{figure}

\paragraph{How accurate is the world model?}
Starting from many states of 10 evaluation episodes, we roll each jet-task world model out autoregressively on the recorded actions and compare the predicted ball positions with the simulator (\cref{fig:wm-analysis}).
Within the 16-step imagination horizon, the coordinate error stays at 2.2--6.2\,mm, about 4--12\% of the 50\,mm ball radius, so the imagined trajectories that the policy is trained on remain close to the real ones.
Beyond the horizon, which the policy never uses, the error grows to 8--19\,mm at 32 steps; the critic covers the return beyond $T$.

\paragraph{What the model captures.}
\Cref{fig:wm-vis} shows a 16-step open-loop rollout on Balance.
The model does not fit the simulator state exactly but learns a low-dimensional embedding oriented toward control: the decoded flow reproduces the overall pattern of the jet but not its absolute speeds or fine details, while the predicted balls stay close to the simulator, which without the fluid would let them fall within a few steps.
It thus captures relative motion and how the jet holds the balls up.

\paragraph{Action ranking.}
Small position errors do not by themselves show that the model can tell better actions from worse ones, which is what policy improvement relies on.
We test this on 32 states from evaluation episodes of each trained policy.
From each state, we record the action sequence that the policy executes over the imagination horizon $T$ and shift it by 0.2 (actions lie in $[-1,1]$) along 15 fixed directions within the action limits; with the unperturbed sequence, this gives 16 candidates.
Every candidate is executed in the simulator, restored from the same full snapshot, and rolled out autoregressively in the world model from the same observation and history, without access to the simulator's future states.
We score the candidates in both rollouts with the same function, rank them by each score, and compute the Spearman correlation between the two rankings, where 1 means identical order and 0 no relation.
As \cref{tab:ranking} shows, the model largely preserves the simulator's ranking of nearby actions.

\begin{table}[t]
\centering\small
\setlength{\tabcolsep}{6pt}
\begin{tabular}{@{}lcccc@{}}
\toprule
 & Balance & Transport & Orbit & Eel \\
\midrule
Spearman $\rho$ & 0.90 & 0.80 & 0.79 & 0.69 \\
\bottomrule
\end{tabular}
\caption{\textbf{Action ranking.} Spearman correlation between the world-model and simulator rankings of 16 candidate action sequences, averaged over 32 states.}
\label{tab:ranking}
\end{table}

\subsection{Task switching}
\label{sec:switching}
A single world model and a task-conditioned policy can also be trained jointly on several jet tasks, and the policy can then switch tasks within an episode without resetting the simulator: it carries two balls once around the Orbit circle, keeps them aloft near their Balance targets, and repeats (see Appendix~\ref{sec:suppl-switching}).

\section{Conclusion}
\label{sec:conclusion}
We presented \ours, an online model-based RL framework that learns a control-oriented world model of fluid--solid interaction jointly with the policy from black-box simulator interactions.
Rather than learning detailed fluid--solid dynamics, the model captures only the low-dimensional dynamics that policy optimization requires, and is refit on the states that the policy visits.
Trained with analytic gradients through imagined rollouts, \ours reaches the final return of the state-of-the-art baseline with several times fewer simulator transitions on the jet tasks, at a fraction of its training GPU time, and attains higher final returns on all four tasks.
The critic is essential: without a value estimate beyond the imagination horizon, the policy fails on Transport.
Choosing what the world model represents, rather than how faithfully it simulates, is what makes it both cheap and sufficient for control. Future work includes real fluidic devices and contact-rich tool manipulation, where the solids respond mainly to contact.

\bibliographystyle{ieeenat_fullname}
{\small\bibliography{main}}

\twocolumn[{\centering\Large\bfseries Supplementary Material\par\vspace{1em}}]
\appendix
\section{Benchmarks}
\label{sec:suppl-details}

\subsection{Tasks and rewards}
\label{sec:suppl-benchmarks}

\paragraph{MPM jet tasks.} In Balance, Transport, and Orbit, a nozzle near the floor shoots a water jet at the balls, and the policy moves the nozzle horizontally, tilts it, and sets the flow rate. The scenes are simulated with MLS-MPM~\cite{hu2018moving} in Warp~\cite{macklin2022warp}; \cref{tab:sim-suppl} lists the settings.

At every step, each ball earns
\begin{equation}
  r = \alpha\big(2.5\,e^{-30 d^2} + 1.5\,e^{-2\|\dot p - \dot p^\star\|}\big),
  \label{eq:suppl-reward}
\end{equation}
where $d$ is a weighted distance from the ball center to its target, $\dot p$ and $\dot p^\star$ are the velocities of the ball and the target, and $\alpha$ is a constant scale per task; the rewards of the balls are summed. If a ball center falls below 0.25\,m, the episode ends and the reward of that step is reduced by $25\alpha$. We keep the targets, weights, and scales of the benchmarks, so the returns are comparable with~\cite{chen2025fast}.

\textit{Balance}~\cite{chen2025fast}: two balls are held at fixed targets 0.20\,m apart at 0.55\,m height, for up to 8{,}000 steps (30\,s). A return of about 100 means that both balls stay close to their targets for the whole episode.

\textit{Transport}: one ball is carried to a sequence of goal cells drawn uniformly from a $3{\times}2{\times}2$ grid of boxes of $0.40{\times}0.20{\times}0.12$\,m; the target $p^\star$ is the cell center. The goal changes every 250 steps (about 1\,s). A goal counts as reached if the ball center enters its cell at any time in the window; otherwise the episode ends at the end of the window. Episodes also end when the ball falls, and otherwise after eight goals (2{,}000 steps). The environment of~\cite{chen2025fast} is not public, so we rebuild it from its description.

\textit{Orbit}: two balls follow antipodal targets that move at 0.55\,m height on a horizontal circle of radius 0.10\,m, one revolution every 8\,s, for up to 8{,}000 steps. The target velocity $\dot p^\star$ is the tangential velocity of the circle (about 0.08\,m/s), so the balls cannot simply settle near the circle.

\paragraph{Eel.} A 12-link eel swims in a 2D lattice-Boltzmann fluid and must follow a given path by bending its 11 joints. Lengths are in lattice cells and time in control steps. An episode fails when the eel is more than 40 cells from the path, when its heading deviates from the path tangent by more than 0.8\,rad during training (1.6\,rad in evaluation, for both methods), or when its body comes within 60 cells of the domain boundary. We test on 25 paths (straight, arcs, sinusoids, random), and a test succeeds if the eel reaches the end of the path within 15 cells of it.

The reward per step is
\begin{equation}
\begin{split}
  r_t = {}& \min(\Delta l_t, \Delta l_{\max})/\Delta l_{\mathrm{ref}} - 0.5\big(1-e^{-(\epsilon_t/10)^2}\big) \\
          & - \big(1-\cos\Delta\psi_t\big) - 5\,\overline{(a_t-a_{t-1})^2},
\end{split}
  \label{eq:suppl-eel-reward}
\end{equation}
where $\Delta l_t$ is the distance the eel advanced along the path during the step, $\Delta l_{\mathrm{ref}}{=}0.12$ and $\Delta l_{\max}{=}0.15$ cells per step are a reference and a maximum advance, $\epsilon_t$ is the distance from the path in cells, $\Delta\psi_t$ the heading error, and the bar averages over the joints. The first term rewards progress, about 1 per step at the reference advance, and the cap keeps the eel from gaining reward by rushing. The other terms keep the eel on the path, pointing along it, and moving smoothly. A small penalty on excessive head speed keeps the flow in the stable range of the solver, reaching the end of the path gives a bonus of 5, and failing a penalty of 5. Since progress dominates, the return tells how far the eel gets along the path, and the success rate whether it gets to the end.

\subsection{Inputs}
\label{sec:suppl-inputs}
\Cref{tab:config-suppl} lists what each network reads. The fluid enters in two ways. The \emph{flow code} (jet tasks) and the \emph{lateral line} (eel) describe the flow right now. The \emph{history} (Eq.~\eqref{eq:history}) records what the bodies and the actuator did in the last few steps. We keep 8 steps on the jet tasks, about the time the water takes to fly from the nozzle to the balls, and 4 steps on the eel. \Cref{sec:suppl-inputs-abl} examines the role of these inputs.

\begin{table*}[t]
\centering\small
\setlength{\tabcolsep}{4.5pt}
\begin{tabular}{@{}lllllccccc@{}}
\toprule
Task & Physical state $q_t$ & Fluid obs.\ $z_t$ & History $h_t$ & Task variables $g_t$ & $d_\pi$ & $d_V$ & $d_{\mathrm{WM}}$ & $k$ & $T$ \\
\midrule
Balance   & balls, nozzle & flow code   & --                & targets         & 90  & 90  & 88  & 8  & 16 \\
Transport & ball, nozzle  & flow code   & 8 steps & goal, clock     & 235 & 235 & 230 & 8  & 16 \\
Orbit     & balls, nozzle & flow code   & 8 steps           & moving targets  & 298 & 298 & 288 & 8  & 16 \\
Eel       & body          & lateral line & 4 steps          & path ahead      & 338 & 338 & 304 & 32 & 32 \\
\bottomrule
\end{tabular}
\caption{\textbf{Inputs per benchmark.} Physical state: positions and velocities of the bodies and the state of the actuator. Flow code: 64-dimensional code of the fluid velocity field from a frozen autoencoder. Lateral line: fluid velocity at 24 points beside the eel's body, in the body frame (48 dimensions). History: previous physical states and executed actions. $d_\pi$, $d_V$, $d_{\mathrm{WM}}$: input dimensions of the policy, the critic, and the world model (state, fluid observation, history, and action; the world model does not read $g_t$). All methods read the same input as our policy. $k$: world-model window; $T$: imagination horizon.}
\label{tab:config-suppl}
\end{table*}

\subsection{Baselines and protocol}
\label{sec:suppl-protocol}

\paragraph{Baselines.} We run the authors' public implementation of Chen~\etal~\cite{chen2025fast}, ported to Warp, with its original hyperparameters, with and without guided exploration (FEDG) on Balance (\cref{sec:suppl-edg}) and without it on the other tasks. SAC follows the soft actor--critic used for fluid control in~\cite{ren2022versatile}, with two Q-functions, target networks updated with $\tau{=}0.005$, batches of 256, a random warm-up phase, and about one update per real transition. The rewards are scaled by a constant factor for learning, while all reported returns use the original reward. Both baselines read the same input as our policy (\cref{tab:config-suppl}).

\paragraph{Flow code.} All methods use the same frozen 3D convolutional encoder of~\cite{chen2025fast}, which compresses the fluid velocity field on the simulation grid into 64 numbers; its architecture and training are described in their supplementary material.

\paragraph{Protocol.} During training, we regularly run 10 deterministic evaluation episodes (25 test paths on the eel) in a separate simulator; they do not count as training transitions. On the jet tasks, the x-axes count the real transitions of all four parallel environments together.

\section{\texorpdfstring{Implementation of \ours}{Implementation of FSI-WM}}
\label{sec:suppl-impl}

\subsection{World-model training}
\label{sec:suppl-wm}

\paragraph{Prediction targets.} As illustrated in Fig.~\ref{fig:method}(b) and defined in Eq.~\eqref{eq:dynamics}, our world model predicts both the physical state and compact fluid observations when the latter are included in the prediction state. For the jet tasks, the fluid observation is a 64-dimensional flow code obtained from a frozen encoder. For the eel, it consists of local fluid velocities measured along the body (the lateral line). Both the physical state and the compact fluid observations are predicted autoregressively during imagined rollouts. The key distinction is that the world model does not evolve the full fluid state, but instead models coupled fluid--solid dynamics in a low-dimensional representation for policy optimization.

\paragraph{Prediction.} As in Eq.~\eqref{eq:dynamics}, the network $\Delta_\omega$ predicts the change of the state $s_t$ over one step from its input $x_t$ (state and history) and the action $a_t$. In practice, its output is scaled by running statistics of the real one-step changes, with mean $\mu_\Delta$ and standard deviation $\sigma_\Delta$ per component,
\begin{equation}
  \hat s_{t+1} = s_t + \mu_\Delta + \sigma_\Delta \odot \Delta_\omega(x_t, a_t),
  \label{eq:suppl-wm-step}
\end{equation}
so that the network outputs values of order one, and $x_t$ is normalized by statistics of the real states. On the jet tasks, these statistics are running estimates over all real transitions collected so far; on the eel, they are fitted once on the recorded gait data (\cref{sec:suppl-schedule}). Quantities that never change, such as the height of the nozzle, are copied instead of predicted. On the eel, the world model additionally predicts the translation and rotation of the body over the step, expressed in the body frame, and composes them with the current global pose to obtain the next pose. This local representation avoids regressing absolute positions.

\paragraph{Loss.} Eq.~\eqref{eq:wmloss} is the jet-task loss. We sample windows of up to $k$ consecutive transitions from individual episodes. The model starts from the first real state and history and then consumes its own predictions; its history is shifted after every step by appending the predicted physical state and the logged action, as in imagination. Every predicted state $\hat s_j$ is compared with the real one $s_j$, component by component, after dividing by the spread $\sigma_i$ of that component in the real data. The components are averaged within groups $\mathcal{G}$ (ball positions, ball velocities, nozzle, flow code) before the groups are summed, so that the 64 numbers of the flow code count as much as the few numbers of the balls, not more. On the jet tasks, all steps after a termination are masked, while the transition into the terminal state is retained, and the loss is averaged over the unmasked steps. On the eel, windows are drawn within single episodes.
On the eel, all components of the body state and the lateral line are averaged together. A second term compares the predicted and real displacement of the body since the start of the window,
\begin{equation}
  \mathcal{L}^{\mathrm{eel}}_{\mathrm{WM}} = \mathcal{L}_{\mathrm{WM}} + \frac{1}{k}\sum_{j=1}^{k}\frac{1}{3}\Big\|\frac{\hat g_j - g_j}{\sigma_g}\Big\|^2,
  \label{eq:suppl-wm-loss}
\end{equation}
where $g_j$ is the position and heading of the body relative to the start of the window, and $\sigma_g$ the spread of its change per step. Because the body moves very smoothly, this spread is tiny and would make the term dominate, so we bound it from below by 0.1 cells for each translation component and 0.02\,rad for the rotation.

\subsection{Training loop and hyperparameters}
\label{sec:suppl-schedule}

\begin{table}[t]
\centering\small
\setlength{\tabcolsep}{4pt}
\begin{tabular}{@{}lll@{}}
\toprule
 & MPM jet tasks & Eel \\
\midrule
Solver & 3D MLS-MPM (Warp) & 2D D2Q9 LBM \\
Grid & $80{\times}48{\times}32$ & $400{\times}600$ \\
Solids & 700 particles/ball & 12 rigid links \\
 & $E{=}800$, $\nu{=}0.2$ & 11 yaw joints \\
Density ratio & 4 & -- \\
Control step & 3.8\,ms & 80\,ms \\
Solver steps$^\ast$ & 19 & 8 \\
Parallel worlds & 4 & 64 \\
Real transitions & 2M & 2.4M \\
\bottomrule
\end{tabular}
\caption{\textbf{Simulation settings.} $^\ast$Solver time steps per control step. Density ratio: ball to water; the jet tasks use 20{,}000 water particles.}
\label{tab:sim-suppl}
\end{table}

\begin{table}[t]
\centering\small
\setlength{\tabcolsep}{3.5pt}
\begin{tabular}{@{}lll@{}}
\toprule
 & MPM jet tasks & Eel \\
\midrule
\multicolumn{3}{@{}l}{\textit{World model}} \\
Hidden & $3{\times}256$ & $3{\times}512$ \\
Window $k$ & 8 & 32 \\
Batch & 256 & 2{,}048 \\
Updates$^\ast$ & 10 & 2 \\
Warm-up updates & 500 & 500 \\
\midrule
\multicolumn{3}{@{}l}{\textit{Policy and critic}} \\
Hidden & $2{\times}256$ / $2{\times}256$ & $3{\times}256$ / $2{\times}256$ \\
Horizon $T$ & 16 & 32 \\
Rollouts per update & 256 & 2{,}048 \\
Policy updates$^\ast$ & 1 & 1 \\
Critic epochs & 5 (4 minibatches) & 2 (4 minibatches) \\
$\gamma$, $\lambda$ & 0.99, 0.95 & 0.99, 0.95 \\
\midrule
\multicolumn{3}{@{}l}{\textit{Optimization and data}} \\
Optimizer & \multicolumn{2}{l}{Adam, $2{\times}10^{-4}$, all networks} \\
Gradient clipping & \multicolumn{2}{l}{norm 1.0} \\
Warm-up data & 5{,}000 random & gait data \\
Replay & last 100{,}000 & last 256{,}000 \\
Exploration & \multicolumn{2}{l}{Gaussian, SD 0.1} \\
\bottomrule
\end{tabular}
\caption{\textbf{Hyperparameters of \ours.} $^\ast$Per 64 real transitions. Warm-up: world-model updates before the policy is trained. Gait data: the recorded runs described in the text.}
\label{tab:hp-suppl}
\end{table}

\Cref{tab:sim-suppl,tab:hp-suppl} list the settings. Throughout training, the world model is continually updated using recent replay data to adapt to the state distribution induced by the evolving policy.

\paragraph{MPM jet tasks.} We first collect transitions with random actions and fit the world model on them. After that, every 64 real transitions are followed by world-model, policy, and critic updates, in this order. The policy optimizer uses $\beta_2{=}0.95$, and the policy gradient is also clipped element-wise at 0.1. Compared with Chen~\etal, who perform one learner update per real transition, we update the policy about 60 times less often, but each update differentiates many imagined rollouts at once.

\paragraph{Eel.} Before training, we let the eel swim with a hand-designed undulating gait and record what happens, while the strength, speed, and turning of the gait change at random and the actions are perturbed with noise. This shows the world model what swimming looks like before the policy can swim. The world model is first trained on these recordings, and they keep providing half of its training data and half of the imagination starts afterwards; they are not counted as training transitions. The baseline starts half of its training episodes from the same recordings. During training, all 64 worlds advance by one step, which gives 64 transitions, and then the networks are updated. The joints can change their targets only gradually, and noise is added to the policy's sensors during training.

\section{Additional results}
\label{sec:suppl-results}

\subsection{Results and cost}
\label{sec:suppl-tables}
On Balance, our policy keeps both balls near their targets for the whole episode in every evaluation, while the episodes of Chen~\etal end early on average. On Transport, our ball never falls and reaches every goal cell, and most of the time it gets close to the center of the cell; the baseline drops the ball in some episodes and reaches the center much less often. On Orbit, our episodes last several times longer than those of the baseline.

\begin{table}[t]
\centering\small
\setlength{\tabcolsep}{4pt}
\begin{tabular}{@{}llccc@{}}
\toprule
Task & GPU & \ours & Chen~\etal & SAC \\
\midrule
Balance   & RTX 3080 Ti & 1.6 & 5.1 & 1.8 \\
Transport & RTX 3090    & 1.8 & 6.6 & 2.3 \\
Orbit     & RTX 4080    & 1.4 & 5.8 & 2.0 \\
Eel       & RTX 3090    & 8.8 & 8.2 & 5.6 \\
\bottomrule
\end{tabular}
\caption{\textbf{Training GPU-hours} to reach 2M real transitions (2.4M on the eel), excluding evaluation. Within each task, all methods are timed on the same GPU. Timings are not tuned for speed.}
\label{tab:cost}
\end{table}

\paragraph{Where the time goes.} Across the three MPM jet tasks, \ours requires 3.2--4.1$\times$ less training time than Chen~\etal for the same number of real transitions (\cref{tab:cost}). The reduction comes mainly from the lower frequency of policy and critic updates and from batched computation over imagined trajectories. On the eel, \ours has a computational cost comparable to Chen~\etal, and its advantage lies in control performance and sample efficiency rather than wall-clock speed.

\subsection{Role of fluid information in model-based control}
\label{sec:suppl-inputs-abl}
Our world model does not simulate the fluid, so it must infer the effect of the fluid on the bodies from its input. These experiments test this design motivation: which network needs information about the fluid, and how much its form matters. We compare two simple forms. The flow code on the jet tasks and the lateral line on the eel observe the current flow directly, whereas a short history of the physical states and executed actions reflects the flow only indirectly.

\begin{figure}[t]
  \centering
  \includegraphics[width=\linewidth]{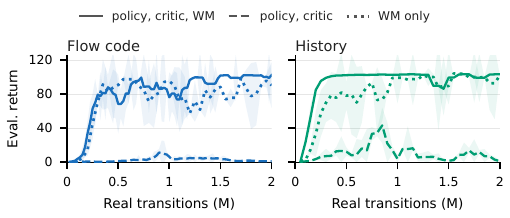}
  \caption{\textbf{Fluid information on Balance.} Left: flow code; right: 8-step history; each given to the policy, critic, and world model (solid), to the policy and critic only (dashed), or to the world model only (dotted). The solid and dashed curves on the left are those of Fig.~\ref{fig:balance-abl}. Lines: moving average; bands: one SD over the 10 evaluation episodes.}
  \label{fig:balance-inputs}
\end{figure}

\paragraph{Balance.} We give the flow code or the history to the policy and the critic only, to the world model only, or to all three networks (\cref{fig:balance-inputs}). Both forms behave the same way. Given only to the policy and the critic, neither makes the task learnable, as already seen for the flow code in Fig.~\ref{fig:balance-abl}. Given only to the world model, both do, even though the policy then sees only the current bodies and targets. Giving the information to all three networks is the most reliable, but most of the gain already comes from the world model.

The policy is optimized on the dynamics that the world model predicts. A world model that cannot tell how the water will hit the balls predicts an average response, and the imagined gradients favor actions that work in this average flow but not in the real one, whatever the policy observes. What matters is therefore that the world model receives enough control-relevant information about the fluid, rather than which form this information takes.

\begin{figure}[t]
  \centering
  \includegraphics[width=\linewidth]{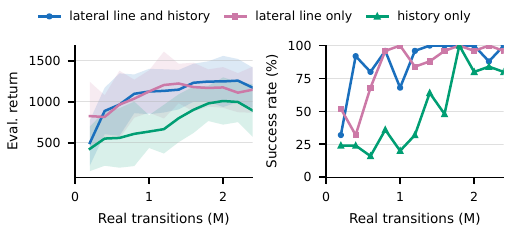}
  \caption{\textbf{Fluid information on the eel.} Lateral line and 4-step history (the run in Section~\ref{sec:main}), lateral line only, or history only, given to all networks. Left: return on the 25 test paths (moving average; bands: one SD over the paths). Right: success rate on the 25 test paths.}
  \label{fig:eel-inputs}
\end{figure}

\paragraph{Eel.} All networks read the lateral line, the 4-step history, or both (\cref{fig:eel-inputs}). On the eel, the direct local flow observation is more informative than the short history alone. Combining the two speeds up early learning, although the lateral line alone reaches comparable performance at convergence. Together with Balance, these results suggest that the suitable representation of the fluid depends on the interaction dynamics. Unless otherwise specified, the world model and the policy receive the same fluid-related observations in our main experiments.

\subsection{Multiple training seeds}
\label{sec:suppl-seeds}
\begin{figure}[t]
  \centering
  \includegraphics[width=\linewidth]{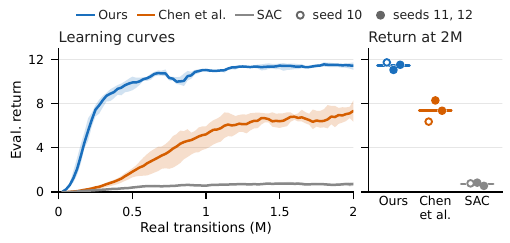}
  \caption{\textbf{Transport over three training seeds.} Left: return vs.\ real transitions; lines: mean over seeds of the moving average, bands: range over seeds. Right: return at 2M of each seed (open: the seed used in Section~\ref{sec:main}) and their mean (bar).}
  \label{fig:seeds-suppl}
\end{figure}

To assess the robustness of our results across random seeds, we repeat the Transport experiments with three training seeds for each method. As shown in \cref{fig:seeds-suppl} and \cref{tab:seeds-suppl}, \ours consistently outperforms both baselines in final return. At 2M real transitions, \ours achieves a mean return of 11.45$\,\pm\,$0.35, compared with 7.35$\,\pm\,$0.96 for Chen~\etal. Moreover, \ours reaches a return of 6 with approximately five times fewer real transitions on average. Although the performance gap varies across seeds, the advantage of \ours is consistent across all three runs. SAC does not reach this threshold within the training budget.

\begin{table}[t]
\centering\small
\setlength{\tabcolsep}{4.5pt}
\begin{tabular}{@{}lcccc@{}}
\toprule
 & Seed 10 & Seed 11 & Seed 12 & Mean \\
\midrule
\multicolumn{5}{@{}l}{\textit{Return at 2M}} \\
\ours      & 11.75 & 11.07 & 11.53 & \textbf{11.45}$\,\pm\,$0.35 \\
Chen~\etal & 6.38  & 8.30  & 7.37  & 7.35$\,\pm\,$0.96 \\
SAC        & 0.78  & 0.84  & 0.55  & 0.73$\,\pm\,$0.15 \\
\midrule
\multicolumn{5}{@{}l}{\textit{Transitions to return $\geq$6 (M)}} \\
\ours      & 0.25 & 0.25 & 0.20 & \textbf{0.23} \\
Chen~\etal & 1.40 & 1.03 & 1.00 & 1.14 \\
SAC        & --   & --   & --   & not reached \\
\bottomrule
\end{tabular}
\caption{\textbf{Transport over three training seeds.} Return at 2M transitions, and transitions until the return, averaged over three consecutive evaluations, first reaches 6. Mean: mean$\,\pm\,$SD over seeds. Seed 10 is the run in Section~\ref{sec:main}.}
\label{tab:seeds-suppl}
\end{table}

\subsection{Shared dynamics and continuous task switching}
\label{sec:suppl-switching}

\paragraph{Joint training.} For Section~\ref{sec:switching}, one world model and one policy are trained together on several jet tasks in the same scene, including two-ball Balance and two-ball Orbit. For this experiment, the scene is made the same in both horizontal directions, the nozzle can move and tilt in both, and the balls are somewhat heavier than in the main benchmarks. All tasks use the reward of \cref{eq:suppl-reward}, with $d$ the plain Euclidean distance. The world model reads the same inputs for every task and does not know which task is running; the task only changes the reward and the termination, which are computed in closed form. The policy is a single network with one set of weights. It is not told the task either: it sees where the targets of the balls are and how they move, and that is what defines the task. Each task has its own critic, because returns differ in scale and length between tasks. The tasks get equal shares of the real data and equal weights in the policy objective. During training, no episode is longer than 3{,}000 steps (11\,s).

\begin{figure}[htbp]
  \centering
  \includegraphics[width=\linewidth]{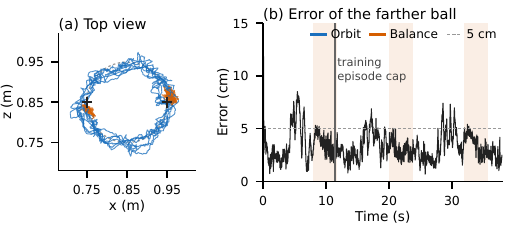}
  \caption{\textbf{Switching between Orbit and Balance without resets} (one of 10 trials). (a) Top view of the ball paths during Orbit (blue) and Balance (orange); dashed: target circle; crosses: Balance targets. (b) Distance of the farther ball from its target; shaded: Balance phases. Vertical line: the longest training episode.}
  \label{fig:switching}
\end{figure}

\paragraph{Continuous task switching.} We evaluate the jointly trained policy by alternating between Orbit and Balance within a single uninterrupted episode, without resetting the simulator or the observation history. At each switch, the new targets start from the previous ones. All ten evaluation trials complete 10{,}000 control steps (38\,s) without either ball falling, exceeding the maximum training episode length by more than three times (\cref{fig:switching}). These results show that the shared policy can accommodate changes in the task objective and sustain control beyond the duration of the training episodes.

\newpage
\subsection{Guided exploration (FEDG) in the parallel port}
\label{sec:suppl-edg}
\begin{figure}[htbp]
  \centering
  \includegraphics[width=0.85\linewidth]{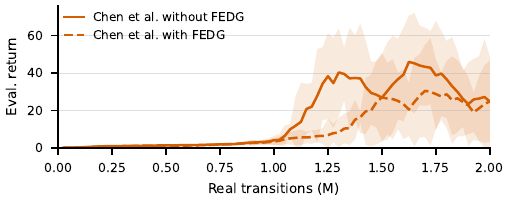}
  \caption{\textbf{Chen~\etal with and without FEDG on Balance.} Return vs.\ real transitions, four parallel environments, same input as in the comparison in Section~\ref{sec:main}.}
  \label{fig:edg-suppl}
\end{figure}

FEDG~\cite{chen2025fast} gives an extra reward when the nozzle is under the lower ball and, part of the time, lets the policy work on this easier auxiliary task. We ran the public implementation with and without FEDG on Balance, with four parallel environments (\cref{fig:edg-suppl}). Both variants end at the same return, and the variant without FEDG peaks higher during training. Under our four-environment training setup, FEDG does not noticeably improve the final evaluation return, while increasing the training time per environment transition by approximately 1.7$\times$ due to additional auxiliary-task updates. We therefore use the variant without FEDG in the main comparisons.

\end{document}